\documentclass[conference, 10pt]{IEEEtran}
\IEEEoverridecommandlockouts
\usepackage{setspace}
\usepackage{graphicx}
\usepackage{epsf,cite,subfigure}
\usepackage{amsmath,amssymb}
\usepackage{floatflt}
\usepackage{url,color}
\usepackage{times}

\newcommand{\E}[1]{E\left[#1\right]}

\newcommand{\set}[1]{\left[#1\right]}

\newcommand{\eqnref}[1]{(\ref{eqn:#1})}

\newcommand{\eqnlabel}[1]{\label{eqn:#1}}

\newcommand     {\paren}[1]{\left(#1\right)}
\newcommand{\abs}[1]{\left|#1\right|}

\newcommand{\eX}[1]{\e^{#1}}

\newcommand{\deriv}[1]{#1^\prime}

\newcommand{\e}{e}

\DeclareMathOperator*{\argmax}{\arg\!\max}

\usepackage{hyperref}

\title{EM-Based Channel Estimation from Crowd-Sourced RSSI Samples Corrupted by Noise and Interference}
\author{
\IEEEauthorblockN{\normalsize Silvija Kokalj-Filipovic\IEEEauthorrefmark{1} and Larry Greenstein}
\IEEEauthorblockA{\small WINLAB, Rutgers University\\
\small\em \{skokalj,ljg\}@winlab.rutgers.edu} 
\thanks{ \IEEEauthorrefmark{1} S. Kokalj-Filipovic is currently affiliated with Naval Research Labs.
}
\thanks{ \IEEEauthorrefmark{2} This version of the paper corrects several typos discovered in the version published in the IEEEXplore.
}
}
\begin{document}
\date{} \maketitle
\begin{abstract}
We propose a method for estimating channel parameters from RSSI measurements and the lost packet count, which can work in the presence of losses due to both interference and signal attenuation below the noise floor. This is especially important in the wireless networks, such as vehicular, where propagation model changes with the density of nodes. The method is based on {\em Stochastic Expectation Maximization}, where the received data is modeled as a mixture of distributions (no/low interference and strong interference), incomplete (censored) due to packet losses. The PDFs in the mixture are Gamma, according to the commonly accepted model for wireless signal and interference power. This approach leverages the loss count as additional information, hence outperforming maximum likelihood estimation, which does not use this information (ML-), for a small number of received RSSI samples. Hence, it allows inexpensive on-line channel estimation from ad-hoc collected data. The method also outperforms ML- on uncensored data mixtures, as ML- assumes that samples are from a single-mode PDF.
\IEEEauthorrefmark{2} \end{abstract}
\section{Introduction} \label{sec:Intro}
For various reasons (such as participatory RF sensing in order to develop low-cost RF maps \cite{GhasemiSousa}, or for calibrating the channel in order to reproduce field trials in a simulator), wireless systems often collect signal strength data “on the fly”, i.e., in the course of actual operation. Such data is often collected in the form of paired values of Tx-Rx distance and the received signal strength indication (RSSI), which can be thought of (within a known additive constant) as the received power in dBm \cite{Vlavianos}.  RSSI is measured on a per-packet basis. If there is too much noise and/or interference for a given measurement, the packet can be lost in which case only the failure indication is recorded (indirectly, e.g., through packet sequencing). The data reduction challenge is to reconstruct, from the collection of recorded RSSI values and packets tagged as lost, the probability density function (PDF) of the received signal. With the PDF thus estimated, the analyst can accurately model the propagation in the environment (e.g., path loss vs. distance), and also model interference effects for a given scenario (e.g., geometry, spatial density of both active and inactive Tx-s, etc.) The widespread adoption of Nakagami  PDFs for modeling radio links  is justified by the abundant analysis of empirical data \cite{Goldsmith,Molisch}. When we refer to the Nakagami PDF, it implies the signal amplitude; the corresponding power is Gamma-distributed, with the same scale parameter m and shape parameter $\Omega$, and the dB power (hence, RSSI) can be thought of as log-Gamma. Note that Nakagami with {\em m=1} corresponds to the Rayleigh distribution.

However, the problem of estimating parameters of this PDF based on packet data collected over time periods of practical interest (the shorter the better) remains challenging. The reason is a high amount of lost (censored) samples caused by interference and low SNR due to fading or distance-based attenuation. As interference is intermittent, there are two broad classes of RSSI data points, namely, those with no (or low) interference, and those with enough interference to result in a significantly modified statistical model (different PDF). Note that maximum-likelihood (ML) \cite{ChengBeaulieu}, the typically best approach for single statistical model, does not offer a closed form solution for data mixtures with loss counts. To derive parameters of PDFs featured in a {\em censored mixture of two random variables} (RVs), representing samples with no/ low interference, and with strong interference, we  propose the use of Stochastic Expectation-Maximization (SEM) \cite{ChauveauSEM} estimators. In addition, our approach leverages the loss count as additional information to improve the estimation accuracy for a given number of samples. We introduce notation {\em ML-} to denote the ML that {\em utilizes a single-mode PDF assumption and only received samples}. In this paper, we demonstrate that our approach performs better than ML- in the presence of interference, because it starts with an assumption of two components (dual mixture) and because it uses the loss count as side information. It also outperforms ML- in cases without interference, if the number of received samples is small, which is frequently the case in on-line estimation tasks.

The organization of the paper is as follows: in Section~\ref{sec:Syst} we briefly describe the system model based on an example, while in Section~\ref{sec:BasicElms}, we introduce basic algorithmic elements; in Section~\ref{sec:ChannelSEM} we present the algorithm used in our approach; we evaluate our model on both simulated and empirical data, and discuss the results in Section~\ref{sec:Eval}. In the last section we conclude and address future work.  
\section{System Model and Motivational Example}\label{sec:Syst}
\begin{figure}[t] 
\begin{center}
\includegraphics[width=2.8in]{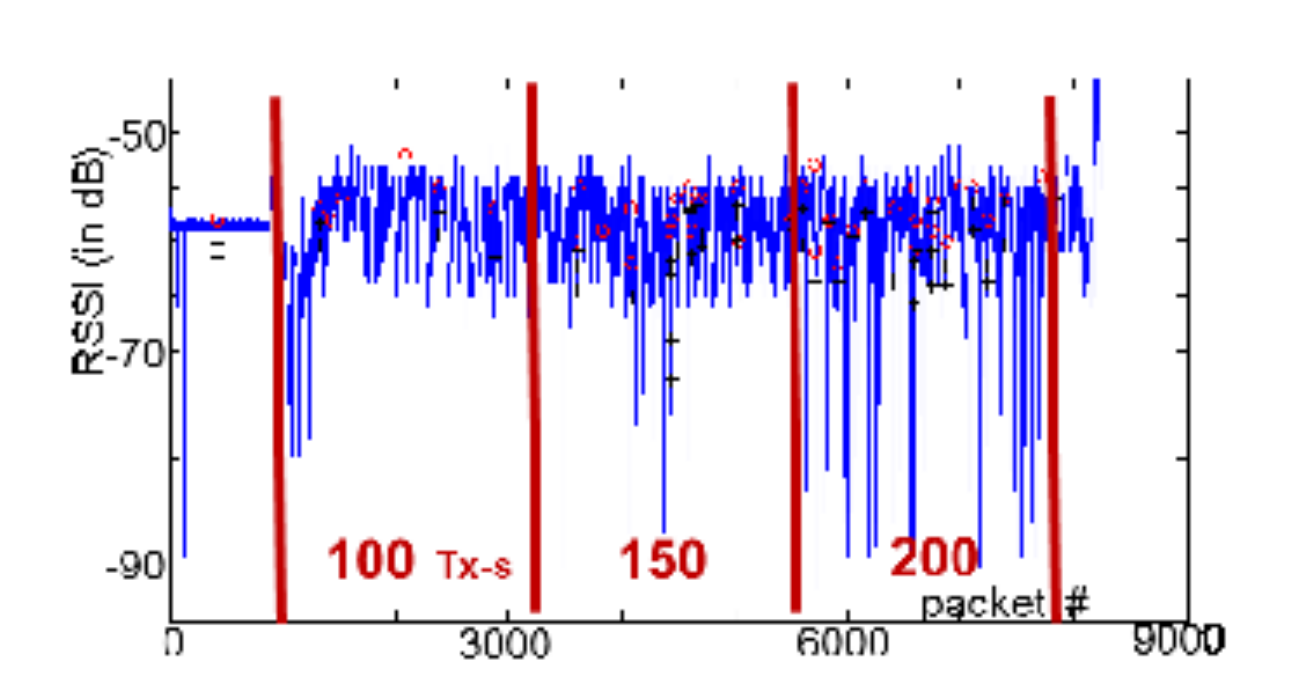}\vspace{-3mm}
\caption{Time plots from \cite{KokaljCamp}, showing the effect of an increasing number of active Tx-s to the RSSI of a single mobile link, with Tx at a constant distance from Rx} \vspace{-5mm}  \label{fig:grad}
\end{center}
\end{figure}

We refer to no/ low interference samples as {\em signal (or 1st) component}, and to strong interference samples as {\em interference (or 2nd) component}. We propose to have both signal and interference components in the mixture modeled by the same family of PDFs, i.e., Gamma. Properly parameterized Gamma PDFs ({\em GPDFs}) are widely used to model small-scale fading, to approximate the product of the small-scale and lognormal fading distribution, and to approximate the interference power \cite{Heath13}. 
Our claim that interference samples deserve to be modeled by a 2nd component is evident in Fig~\ref{fig:grad} \cite{KokaljCamp}, where the distortion caused by interference increases with the spatial density of interferers. The field trial in which the samples were collected included 200 (moving) vehicles equipped with wireless modems, where the test first ran with 100 active transmitters; then another 50 were added, and in the final third of the test all 200 modems were transmitting (3 parts delineated in Fig~\ref{fig:grad}). Note that these and other RSSI measurements  featured here are made on OFDM transmissions with a 10MHz bandwidth centered near 5.9 GHz, in compliance with V2V DSRC IEEE802.11p, using Atheros 802.11p chips.
It appears that fading is increased as more Tx-s are activated in the field, although the propagation environment has not changed, due to constant density of vehicles. This is the effect of random phases of the interferers; the sum of the M random phasors with equal amplitudes approaches Rayleigh as M grows. Hence, as the interference increases, m in Gamma (and Nakagami, for amplitudes) should approach 1. In this case, the dB peak power (as in Fig~\ref{fig:grad}) is limited to be 10log10(M) above the average power, but the dB power swings below the average can be huge, because of the phasor-sum reductions. In the Rayleigh limit (which M = 10 roughly approximates), the probability of being 10 dB or more below the average is about 10\%, while the prob. of 10 dB or more above the average is 0. 

For this reason, we would model the 2nd component in the mixture with a GPDF of the scale parameter m initially set to one, while the 1st component (pure signal) is modeled with a different m, initially set according to some side information about the data origin (mobile, static, indoor, outdoor, rural, urban etc). Starting with these and other initial values, the SEM algorithm should eventually converge to parameters that better characterize both the signal and the interference as functions of the distance from the signal Tx. In each RSSI mixture component, there are two sub-classes: received (uncensored) data, and lost (censored) data. For the no/low interference case, the censored data are mostly at large distances where the median Rx power is attenuated at or below the noise threshold. The Rx power can also go below the noise floor at any distance as a result of deep fades, due to multi-path. Per Fig~\ref{fig:grad}, the interference causes similar fading on RSSI samples, possibly more intense, causing more losses. 

The stochastic EM algorithm is a known approach for computing ML estimates in the mixture problem. Our model is derived from an extension of the SEM algorithm \cite{CeleuxDiebolt}, dubbed SEMcm,  in a particular case of incomplete data \cite{ChauveauSEM}, where the information loss is due to both mixture of distributions and censored observations. We aim to estimate the parameters of a left-censored dual mixture, which we propose as a model of observed wireless RSSI samples with countable losses, following \cite{ChauveauSEM}. 
\section{Basic Algorithmic Elements}\label{sec:BasicElms}
\begin{figure}[t] 
\begin{center}
\includegraphics[width=3.6in]{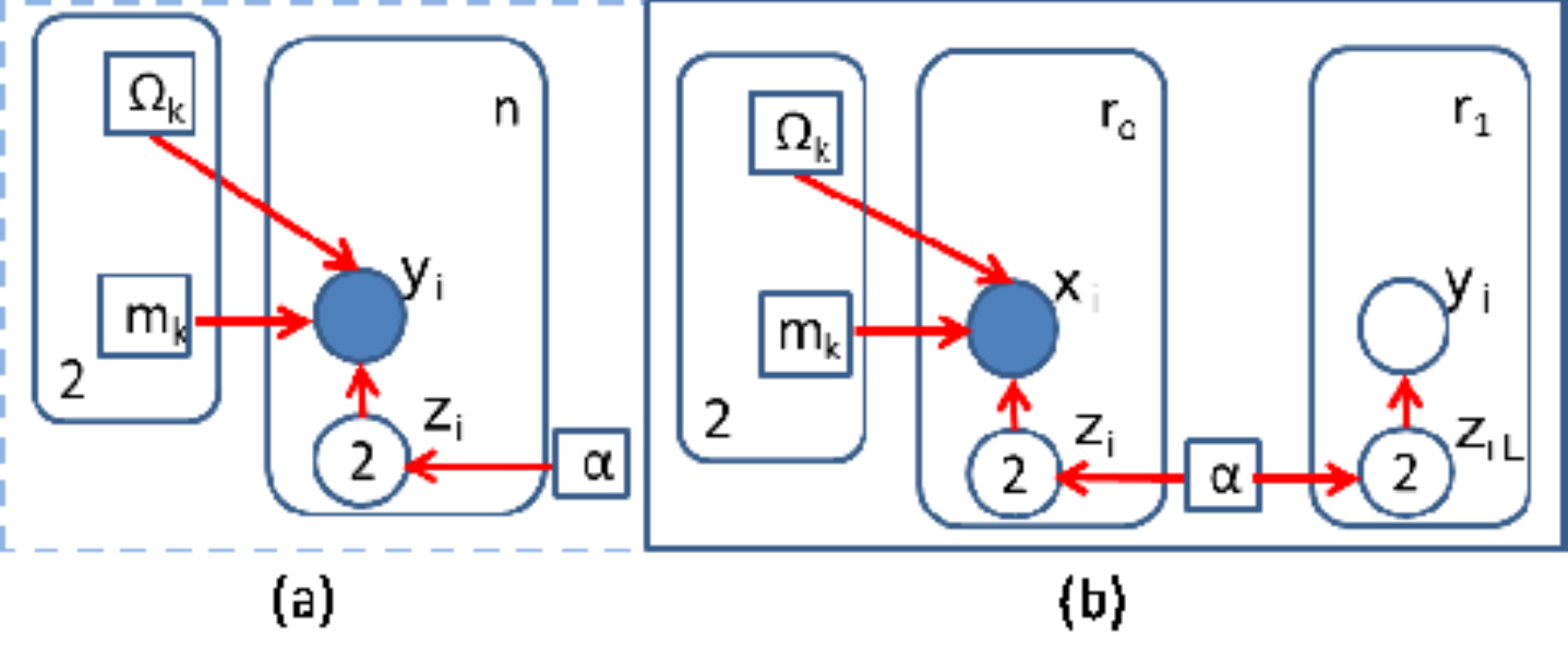}\vspace{-3mm}
\caption{Plate models for (a) uncensured dual mixture of Gamma components (b) censured dual mixture of Gamma components; Shaded circles represent observables.} \vspace{-5mm}  \label{fig:plate}
\end{center}
\end{figure}

A mixture of 2 distributions of the same family $p(y|\phi_i),\ i=1,2,$ is defined by
\begin{align}\eqnlabel{mixt1}
p_{\varphi}(y)=\alpha_1p(y|\phi_1)+\alpha_2p(y|\phi_2).
\end{align}
Here, y is the RV modeling an arbitrary mixture sample.
$\alpha_1$ is the mixing probability. Equivalently, 
\begin{align}\eqnlabel{mixt1a}
p_{\varphi}(y,z)=p(y|\varphi_z)= f_{\varphi_z}(y)
\end{align} 
is the joint distribution of the RVs Y and Z,  where Z is the indicator RV modeling the association with one of the two mixture components (with probability $\alpha_i$), and the subscript  represents the PDF parameters that we aim to estimate:
\begin{align}\eqnlabel{mixt2}
\varphi=\paren{\alpha_1,\alpha_2,\phi_1,\phi_2},\ 0\leq \alpha_1 \leq 1,\ \alpha_1=1-\alpha_2.
\end{align}
We propose that de-logged RSSI values be modeled by a dual mixture of GPDFs $p(y|\phi_i),\ i=1,2.$ Hence, we have  
\begin{align}\eqnlabel{mixt3}
\phi_i=\paren{m_i, \Omega_i};\ p(y|\phi_i)=\frac{1}{\Gamma(m_i)\Omega_i}\paren{\frac{y}{\Omega_i}}^{m_i-1}\eX{\frac{y}{\Omega_i}}.
\end{align}
This model is also depicted in plate notation in Fig.~\ref{fig:plate}~(a). 

Next, we introduce censoring: let $y\in R$  where R is partitioned into disjoint domains $R=R_o\cup R_1,$ where $R_o$ is the subset of uncensored data, wile $R_1$ is the subset corresponding to left-censored data, i.e., $y\le c_L$ where $c_L$ denotes left threshold. Let us assume that there are n samples total (e.g., n transmitted packets), $r_o$ of which are uncensored (received packets): $y_k =x_k \in R_o, k \in C_o, \abs{C_o}=r_o,$ and $r_1$ left censored (lost) samples: $y_k \in R_1, k \in C_1, \abs{C_1}=r_1,$ where $r_o+r_1=n$.
Note that $C_o$ and $C_1$ are disjoint sets of sample indices (e.g., packet sequence numbers SNs), $x_k$ is measured while $y_k$ is te real value (which are not equal for censored samples). In our model, total number od samples and losses could be obtained by tracking SNs of received packets. We define
\begin{align}\eqnlabel{mixt4}
T^{\paren{p+1}}_{o,i}(x_k)=\E{Z_{k,i}|y=x_k,\phi^{(p)}}=\frac{\alpha^{(p)}_i p(x_k|\phi^{(p)}_i)}{\sum^{2}_{t=1}{\alpha^{(p)}_t p(x_k|\phi^{(p)}_t)}},
\end{align} 
where $i=1,2,\ k\in C_o,$ $T^{\paren{p+1}}_{o,i}(x_k)$   denoting {\em current estimate} of the probability that uncensored sample $x_k$ belong to component $i$; and
\begin{align}\eqnlabel{mixt5}
T^{\paren{p+1}}_{1,i}=\E{Z_{i
L}|y\in R_1,\phi^{(p)}}=\frac{\alpha^{(p)}_i \int_{R_1}{p(y|\phi^{(p)}_i)dy}}{\sum^{2}_{t=1}{\alpha^{(p)}_t \int_{R_1}{p(y|\phi^{(p)}_t)dy}}},
\end{align} 
where $i=1,2,$ $T^{\paren{p+1}}_{1,i}$  denoting {\em current estimate} of the probability that a left-censored sample belong to component $i.$ 
The current estimate refers to the $(p+1)$th iteration of the {\em $SEM_{cm}$} algorithm (described in the next subsection). Observe that we have 2 classes of binary latent variables in \eqnref{mixt4} and \eqnref{mixt5}, for $k\in C_o$  and $k\in C_1,$ respectively. The $1$st includes $r_o$  indicators $Z_{k,1}$ characterized by “prob. of success“ $T^{\paren{p+1}}_{o,1}(x_k)$ (prob. of the 1st component),  with $Z_{k,1}=1-Z_{k,2}$; the $2$nd class has a single RV $Z_{1L}$  indicating the 1st component w.p. $T^{\paren{p+1}}_{1,1},$ with $Z_{1L}=1-Z_{2L}.$
The censored model is also depicted in plate notation in Fig.~\ref{fig:plate}~(b). 
\vspace{-2mm} 
\section{SEM-based Channel Estimation Algorithm}\label{sec:ChannelSEM}
Given samples of RSSI, and loss counts for different distances $d$ between a Tx and an Rx, the goal is now to obtain $\alpha_i$ and the two PDFs, $p(y|\phi_i),$ for $i=1$ (signal component) and $i=2$ (interference component), as a function of distance $d.$ We refer to all lost samples as left-censored, as the noise floor is on the left side of the support set of both components, and to the noise floor as the left threshold $c_L.$ Let us first revisit the EM algorithm for mixture data without censoring. We have samples $y,$ but we are missing the indicator RVs $z$ in \eqnref{mixt1a}. The EM algorithm replaces the maximization of the unknown $\log{p_{\varphi}(y,z)}$ by iterative maximizations of the log-likelihood expectation, conditionally to the observed sample $x$, and for the current value of the parameter $\varphi$ \cite{SheskinNonParam}.

To calculate $Q(\varphi,\varphi^{(p)})=\E{\log{p_{\varphi}(y,z)|y=x,\varphi^{(p)}}}$ we must derive the current conditional density of $(y,z)$  given $y=x,$
\begin{align}\eqnlabel{mixt6}
h(y,z|y,\varphi^{(p)})=\frac{p_{\varphi^{(p)}}(y,z)}{f_{\varphi^{(p)}}(y)}.
\end{align} 
Iteration  p+1 has 2 steps:

\noindent {\bf E-step:} Compute $h(y,z|y,\varphi^{(p)})$   (hence $Q(\varphi,\varphi^{(p)})$)

\noindent {\bf M-step:} Choose  $\varphi^{(p+1)}=\argmax_{\varphi\in\Phi}Q(\varphi,\varphi^{(p)}).$

Now, the stochastic EM (SEM) was introduced \cite{CeleuxDiebolt} to overcome the numerical limitations of EM. For the current value $\varphi^{(p)}$ of the parameter, it completes the observed samples by replacing each missing data by a value drawn at random from $h(y,z|y,\varphi^{(p)})$  (S-step), and then computes the ML estimate based on the completed sample (M-step).  We first define the three steps for the left-censored dual-mixture in general, and then present the specific expressions for GPDF.

{\bf E-step:} Compute $T^{\paren{p+1}}_{o,i}(x_k)$ for $k\in C_o,\ i=1,2$

\noindent \qquad\qquad Compute $T^{\paren{p+1}}_{1,i}$ for $i=1,2$

{\bf S-step: (1)} For $x_k \in R_o, k \in C_o$ simulate $r_o$  binary vectors $z^{(p+1)}_k=\set{z^{(p+1)}_{k1},z^{(p+1)}_{k2}}$  by running Bernoulli experiments w.p. $T^{\paren{p+1}}_{o,1}$ ; {\bf(2)} simulate $r_1$ binary vectors $z^{(p+1)}_{Li}=\set{z^{(p+1)}_{Li1},z^{(p+1)}_{Li2}},$  $i=1,\cdots, r_1,$ each as a Bernoulli experiment w.p. $T^{\paren{p+1}}_{1,1}$; {\bf(3)} simulate $r_1$ missing left censored values sampling from  $h(\cdot|c_L,\varphi^{(p)})=\frac{p_{\varphi^{(p)}}(\cdot)}{\int_{R_1}{f_{\varphi^{(p)}}(y)}dy}$; 
\noindent\begin{align}\eqnlabel{mixt7}
\mbox{{\bf M-step:} Choose  }\varphi^{(p+1)}=\argmax_{\varphi\in\Phi}Q(\varphi,\varphi^{(p)})
\end{align} 
where
\begin{align}\eqnlabel{mixt8}
\nonumber &Q(\varphi,\varphi^{(p)})=\sum^{2}_{i=1}{\paren{\sum_{k\in C_o}{z^{(p+1)}_{ki}}+\sum^{r_1}_{j=1}{z^{(p+1)}_{Li,j}}}}\log\alpha^{p}_i+\\
&\sum^{2}_{i=1}{\paren{\sum_{k\in C_o}{z^{(p+1)}_{ki}\log p(x_k|\phi^p_i)}+\sum^{r_1}_{j=1}{z^{(p+1)}_{Li,j}\log p(y_{L,j}|\phi^p_i)}}}
\end{align} 
We next evaluate $Q(\varphi,\varphi^{(p)})$ for GPDFs, resulting in the proposed channel estimation algorithm, dubbed {\em SEMcmG}:

\noindent {\bf E-step:} as in \eqnref{mixt7}-E, based on \eqnref{mixt3}

\noindent {\bf S-step:} as in \eqnref{mixt7}-S, do {\bf (1)-(3)}, based on \eqnref{mixt3}

\noindent {\bf M-step:} Based on  \eqnref{mixt3} and \eqnref{mixt8} solve
\begin{align}\eqnlabel{mixt9}
\nonumber &\mbox{i.  }\frac{\partial Q(\varphi,\varphi^{(p)})}{\partial\alpha_i}=0\Rightarrow\alpha^{(p+1)}_i=\frac{1}{n}\paren{\sum_{k\in C_o}{z^{(p+1)}_{ki}}+\sum^{r_1}_{j=1}{z^{(p+1)}_{Li,j}}},\\
\nonumber &\mbox{ii.  }\frac{\partial Q(\varphi,\varphi^{(p)})}{\partial\Omega_i}=0\Rightarrow \Omega^{(p+1)}_{i}=\frac{\Omega^{(p+1)}_{im}}{m^p_i}\\
\nonumber &\Omega^{(p+1)}_{im}=\frac{\sum_{k\in C_o}{z^{(p+1)}_{ki}x_k}+\sum^{r_1}_{j=1}{z^{(p+1)}_{Li,j}y^{(p+1)}_{L,j}}}{n\alpha^{(p+1)}_i}\\
\nonumber&\mbox{iii.  }\frac{\partial Q(\varphi,\varphi^{(p)})}{\partial m_i}=0\Rightarrow\Psi(x)=\frac{\deriv{\Gamma}(x)}{\Gamma(x)}\\
\nonumber&\Psi(x)\approx \log{x}-\frac{1}{2x}-\frac{1}{12x^2};\ L^{p+1}_{i,x}=\log{\frac{x}{\Omega^{(p+1)}_{i}}}\\
&L^{p+1}_{iA}=\frac{\sum_{k\in C_o}{z^{(p+1)}_{ki}L^{p+1}_{i,x_k}}+\sum^{r_1}_{j=1}{z^{(p+1)}_{Li,j}L^{p+1}_{i,y^{p+1}_{Lj}}}}{n\alpha^{(p+1)}_i}
\end{align} 
$$\mbox{Solve  }\Psi(m^{p+1}_i)-L^{p+1}_{iA}=0.$$

Note that we are frequently averaging over the expected number of samples. Total number of samples and losses could be obtained by tracking sequence numbers of received packets. 
\section{Evaluation} \label{sec:Eval}
\subsection{Model Evaluation on Simulated Data}
\begin{figure}[t] 
\begin{center}
\includegraphics[width=3.5in]{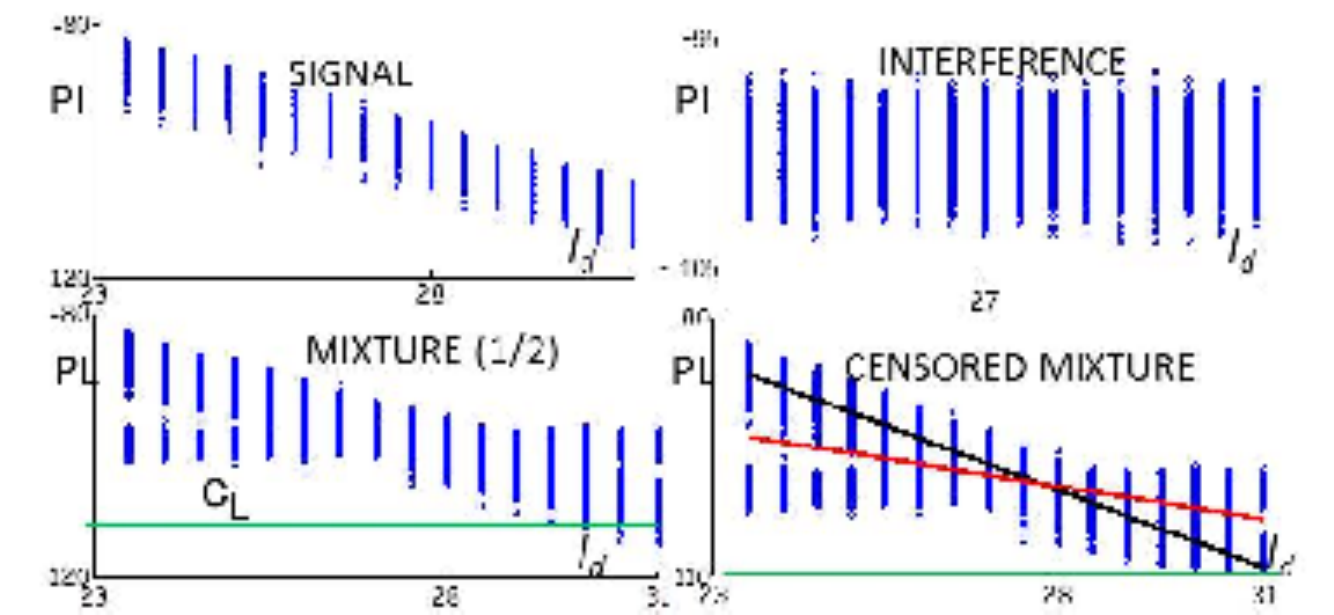}\vspace{-1mm}
\caption{The {\bf making of} of left-censored 2-mixture representing a mobile Rx signals (PL) with interference and attenuation} \vspace{-5mm}  \label{fig:making}
\end{center}
\end{figure}

Besides evaluating SEMcm algorithm on some trivial data sets (one component with left-censoring \cite{GaussMixtWiFi}; one doubly-censored component), we successfully evaluated SEMcmG on a simulated mixture of two left-censored components, which was meant to emulate interference affected RSSI samples. The first component represents the signal over a distance range identical to the range considered in the empirical data evaluation: $l_d=23--32$, where $l_d$ is the log-distance, defined as $10\log{}_{10}(\mbox{distance in m})$. The second component models RSSI samples with strong interference over the same distance range. We simulated different parameters, mostly with the interference component having m=1 (i.e., $m_2\approx 1$), following our discussion in Section~\ref{sec:Intro}. The results are encouraging. However, we now present a mixture with arbitrary parameters, chosen to create a ”signal cloud” visually distinguishable from the interference cloud in the mixture scatter-plot (bottom left pane in Fig.~\ref{fig:making}), while capable of exemplifying main concerns about censored RSSI mixtures. The $m_1$ is chosen slightly high for the assumed mobile signal $(m_1=7),$ while $m_2 =35;$ such a high value of $m_2$ may represent a single (or dominant) interferer. 

Signal attenuation over space is exponential, with the attenuation coefficient to be determined through parameter estimation. We choose to present the exponential attenuation in dB domain as a linear function of $l_d.$ Hence, as in our prior work \cite{KokaljCamp}, median path-loss [PL] is fitted by the straight-line function
\begin{align} \eqnlabel{pl}
 \set{PL}=A-Bl_d.
\end{align}                                                   
Note that PL is defined as $PL=RSSI-10\log{}_{10}(P_t)$, where $P_t$ is the Tx power. Hence, it is distributed as log-Gamma. We present data points in some of our plots as PL rather than RSSI, as it reflects the propagation medium only (independent of Tx power). The simulated $\Omega$ was chosen so that the linear fit into the dBm value of the Gamma mean (i.e, $10\log{}_{10}(\Omega m)$) vs. Tx-Rx distance be equal to \eqnref{pl} with A=-16, B=3. Note that $\Omega$ is a function $\Omega(l_d).$  With these values, the signal only scatter plot (in dB) is presented in the upper-left corner of  Fig.~\ref{fig:making}. 
As for interference, for simplicity and without loss of generality, we propose that the median interference is constant over space, e.g., assuming one distant interferer. Such interference points (dBm) are shown in the upper-right plot in Fig.~\ref{fig:making}. 

Notice that for both components we generated points for discrete values of $l_d,$   referred to as distance bins, with 0.5 dB space in between. For each bin we generated 1000 signal (or interference) points, referred to bin arrays. Then, for each bin, and each bin’s sample index (1-1000) we would select with probability ½ either signal or interference  point in that place, making a balanced mixture of the two components, and ending up with 1000 points per bin (bottom-left in Fig.~\ref{fig:making}). The choice of the mixing coefficient ½ that gives equal weight to both components is deliberate, as such mixtures were hardest to separate. Finally, we censor (drop) the points that are below the threshold cL = -109 (indicated in bottom plots of Fig. 3), resulting in a set of points in the bottom-right plot of Fig.~\ref{fig:making}. These are the points fed into SEMcmG, along with the initial values of the parameters, and the information of how many samples per bin were censored. The initial values were distorted with respect to the real values up to 50
\begin{figure}[t] 
\begin{center}
\includegraphics[width=3.6in]{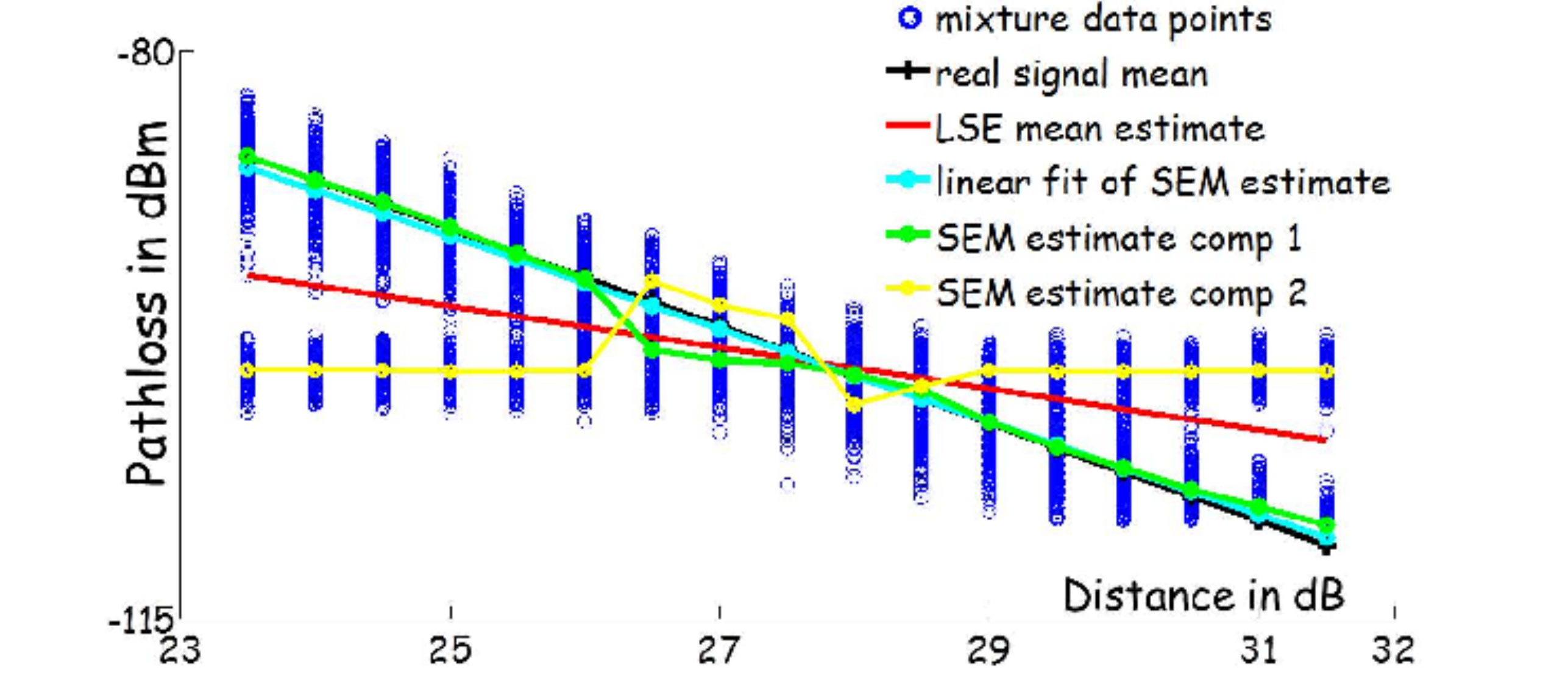}\vspace{-3mm}
\caption{Estimated mean (green) is almost identical to the real one for most distances (except for cluster-overlap bins), so that its linear fit (cyan) is covering the black line (real mean from the bottom-right plot of Fig.~\ref{fig:making}). } \vspace{-5mm}  \label{fig:InterfMean}
\end{center}
\end{figure}
\begin{figure}[t] 
\begin{center}
\includegraphics[width=3.6in]{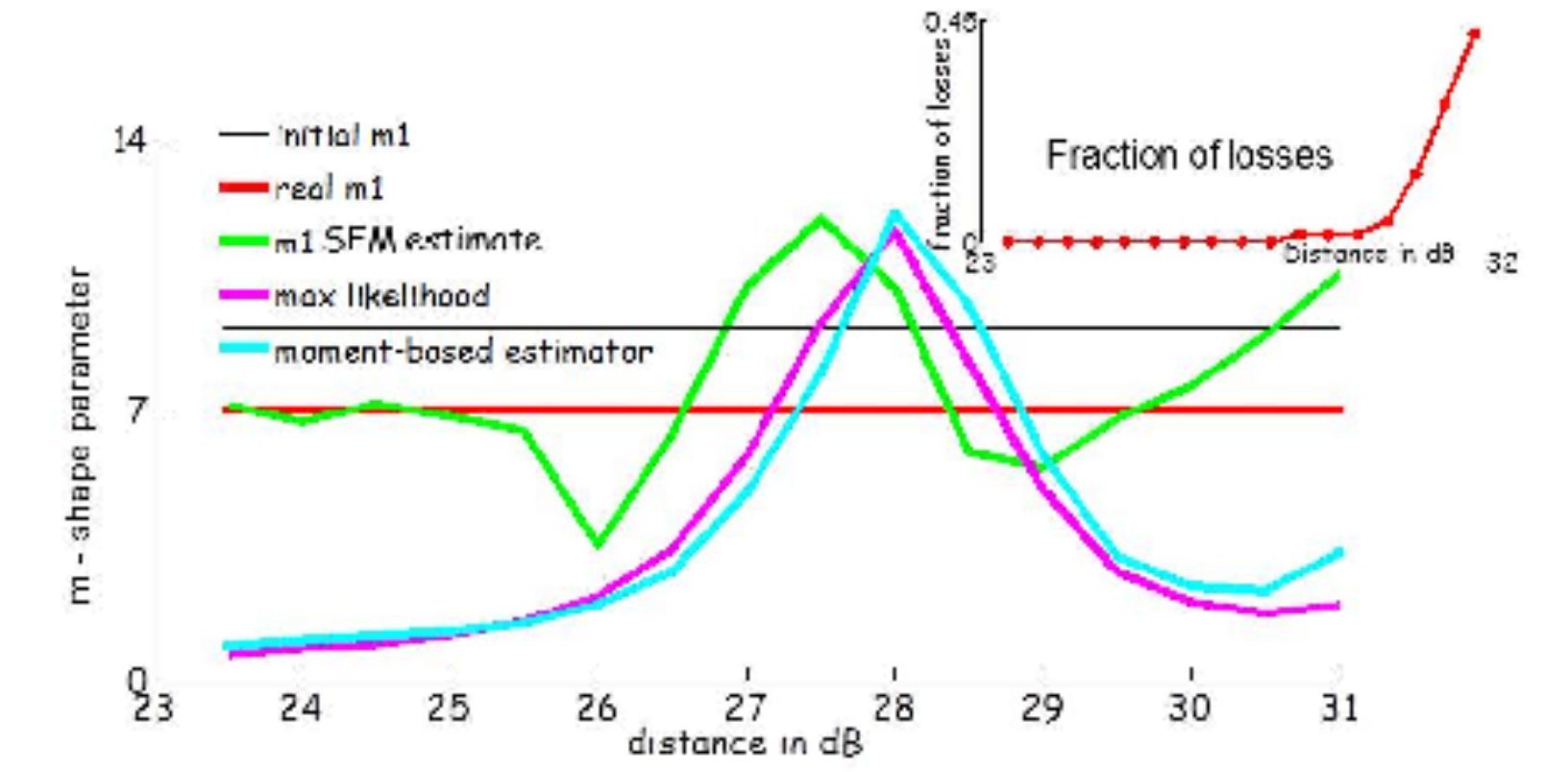}\vspace{-3mm}
\caption{m parameter’s SEM estimate diverges from the real mean in cluster-overlap bins (as do ML and MB).  For other bins, ML and MB take the interference as part of the signal and estimate higher fading (m below 1).} \vspace{-5mm}  \label{fig:mixtM}
\end{center}
\end{figure}
Observe in the bottom-right plot of Fig.~\ref{fig:making} the red line that was obtained as a Least Square Error (LSE) estimate of the mean of the censored mixture, as opposed to the black line that represents the real mean of the signal. This illustrates how much the assumption of one component (as in the presented LSE) can cost in terms of estimation error.  With SEM, the estimates (per bin) were perfect for most simulated mixtures if the data losses constituted less than 60-70\% of data, while for higher losses they were just better than ML estimates.  For this particular mixture, losses were up to 45\% (Fig.~\ref{fig:mixtM}), in order to highlight the “cluster overlap” problem, i.e. the distance bins where the median values of the components were indistinguishable.  Please observe the green line in Fig.~\ref{fig:InterfMean}, which illustrates the signal’s mean estimate. Note that only in the area around $l_d =27$  (cluster overlap) does SEM diverge from the real mean, while following the LSE mean estimate, and in the same area the interference mean estimate follows that of the real signal. We are looking into additional mechanisms to address this phenomenon. 

 Fig.~\ref{fig:mixtM} shows the $m$ estimate per bin. Again, as the likelihood equations are intractable for any maximum likelihood estimate, we compare our results for the $m$ parameter with good existing approximations. The {\em ML and moment-based (MB) } estimates in Fig.~\ref{fig:mixtM} are calculated based on the $r$ received samples. The former one is obtained according to the following maximum likelihood approximation $$m^{ML}=\frac{6+\sqrt{36+48\Delta}}{24\Delta},\mbox{ \cite{ChengBeaulieu}, where}$$ 
$$\Delta=\ln{\set{\sum^{r}_{i=1}{p_i}}}-\frac{1}{r}\sum^{r}_{i=1}{\ln{p_i}}$$ and $p_i$  is the Rx power sample (de-logged RSSI). The latter, $m^{MB},$ follows eqn. (10) from \cite{AbdiKaveh}, which is based on the first two sample moments of the received power $p_i$.
The ML and MB estimates never outperform the SEM estimate, even not in the “cluster overlap” area (Fig.~\ref{fig:mixtM}).  In fact, outside the “overlap”, ML and MB are producing huge errors, as they assume one Gamma distribution, and, hence, they are interpreting the wide clouds outside of central area as a sign of deep fades; consequently, $m$ is estimated too low (around 1). This is a very important argument for the proposed approach, as interference clearly cannot be accounted for by any single-component model. 
\begin{figure}[t] 
\begin{center}
\includegraphics[width=3.0in]{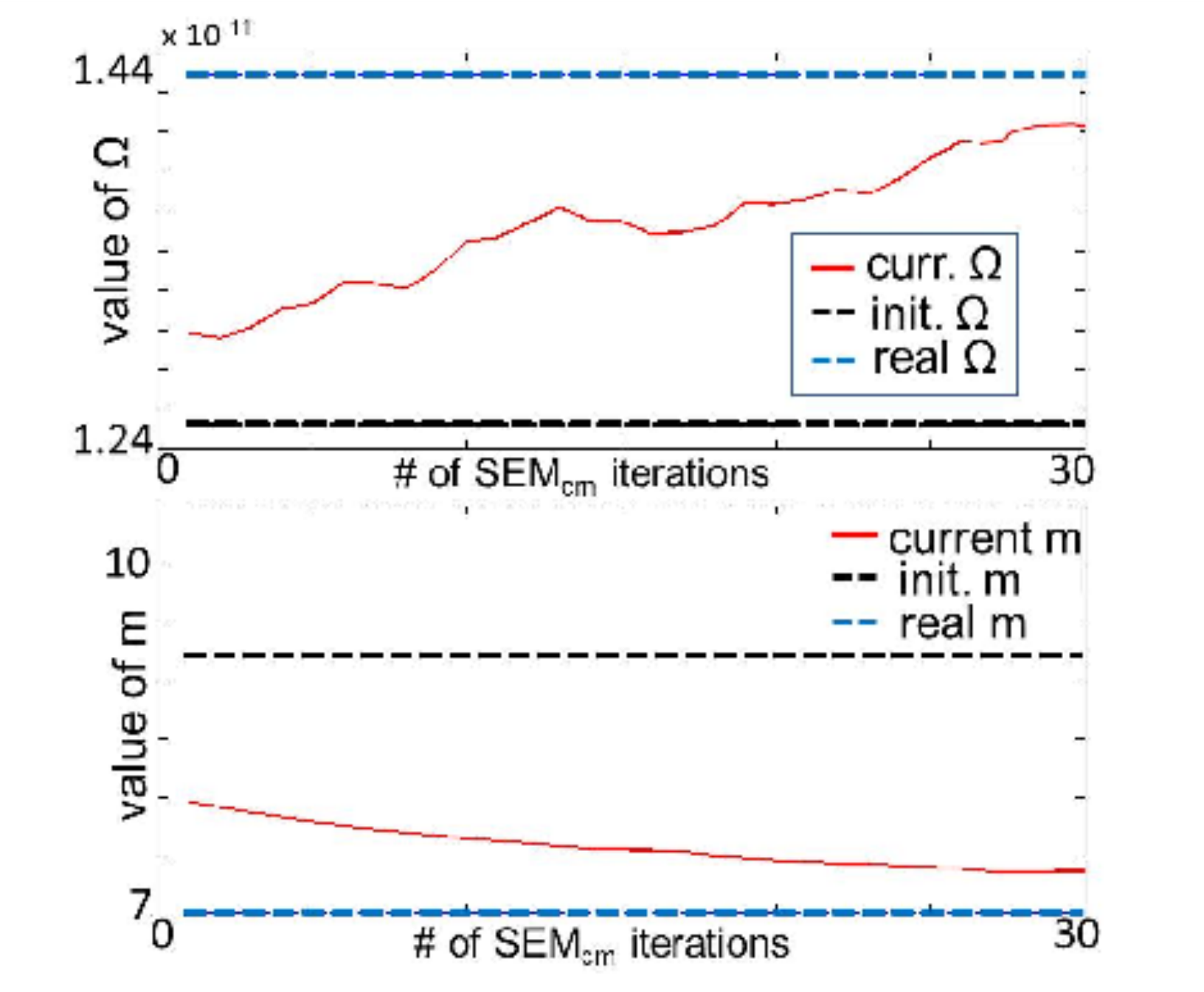}\vspace{-3mm}
\caption{Parameter estimates convergence over 30 iterations to known real values (bin 10):  $\Omega$ - upper plot, m estimate - bottom.} \vspace{-5mm}  \label{fig:both}
\end{center}
\end{figure}
A feature of interest for on-line estimation is the convergence rate.  We illustrate it in Fig.~\ref{fig:both} for both $\Omega$ and $m$ and a given bin: a step-by-step evolution of the estimated parameter. It seems that both estimates could have been better if we ran some additional iterations.
\subsection{Model Evaluation on Empirical  Data}
\begin{figure}[t] 
\begin{center}
\includegraphics[width=3.5in]{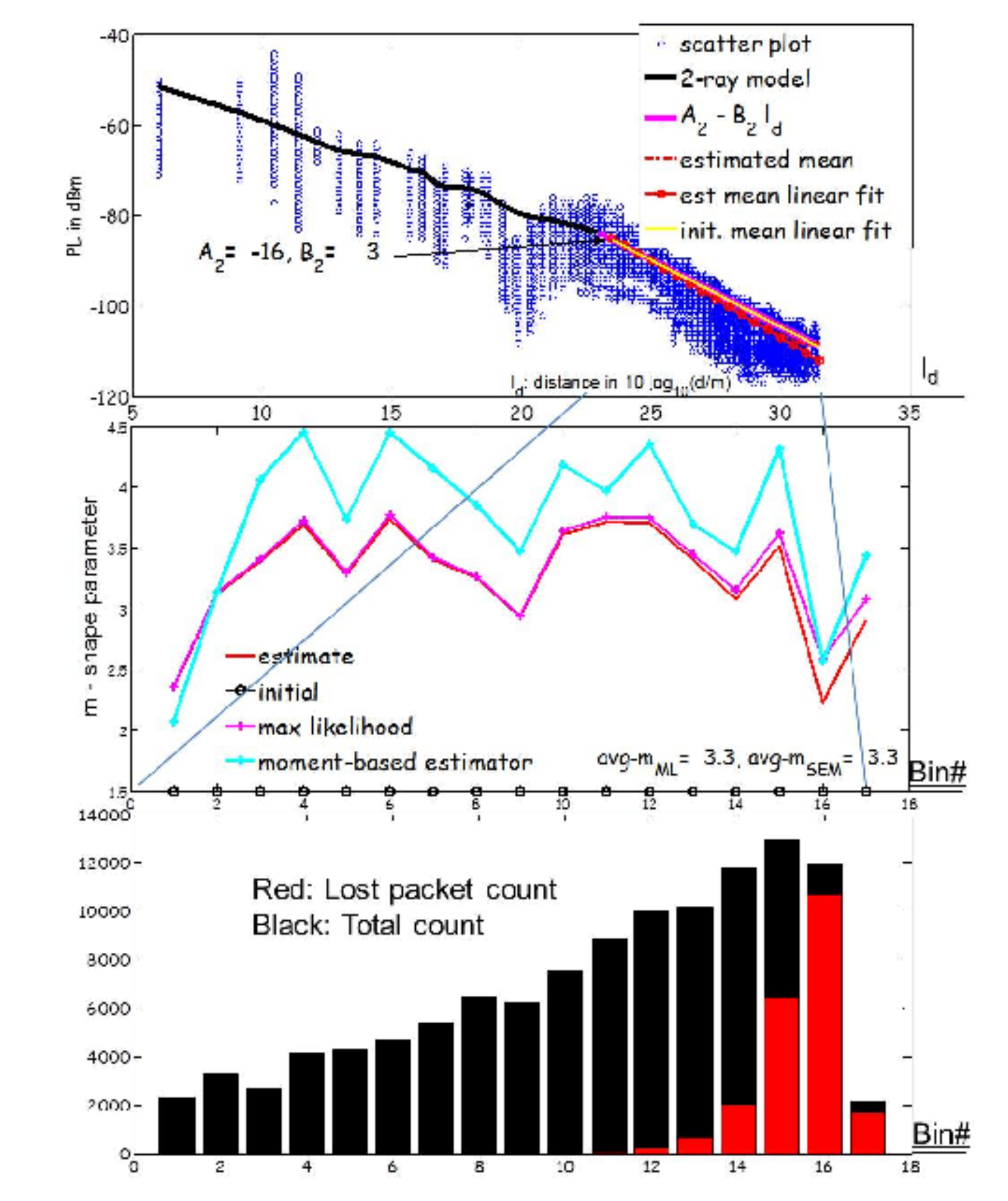}\vspace{-3mm}
\caption{Both mean $(10\log{}(\Omega m))$ in the top plot, and $m,$ in the middle, follow ML estimates when losses are  < 60\%  (bottom).} \vspace{-3mm}  \label{fig:ZRpt18}
\end{center}
\end{figure}

The ZR trial, described in \cite{KokaljCamp}, included only one Tx at a time, mounted on a vehicle that traveled back and forth from the static Rx-s on a straight road $d_{max}$=1200 m long. This scenario with no interference helped us to study the performance of our SEMcm algorithms in terms of signal component estimates, when the initial values for the (non-existent) interference were arbitrary. As the Tx was mobile, suggesting Gamma distribution, SEMcm with Gaussian PDFs gave bad estimates (as expected) and numerical instabilities. SEMcmG showed good results. The initial values for the signal parameters were taken from imperfect estimates, based on the linear LSE fit into a pathloss function that was linear only beyond a break point, and also due to noise-floor saturation (Fig.~\ref{fig:ZRpt18} – upper pane). 

For simplicity we performed SEMcmG only for the distance bins after the break-point (2nd segment), as the smaller distances involved the two-ray phenomenon. The linear fit of the initial $\Omega m$ in dBm, represented by the yellow line in Fig.~\ref{fig:ZRpt18}, matches \eqnref{pl} with coefficients A2 and B2.  Other coefficients, based on the LSE over the 2nd segment only, came closer to the real median PL (known from running the same field trial with higher Tx power, which avoids the noise floor within traversed distances). 

The SEMcm estimated line (red line with circular markers) is almost the same as the real one. The initial value for m was 1.5 (bottom black line in the middle pane of Fig.~\ref{fig:ZRpt18}), yet SEMcmG managed to improve it to 3.3 on average, which is identical to its ML estimate. Now, the ML estimate works optimally when there is sufficient number of samples, which was the case here.  The bottom pane of Fig.~\ref{fig:ZRpt18} shows the number of transmitted packets in black, and the number of lost packets in red. The last bin has the worst losses (75\%), yet, more than 500 packets received is sufficient for ML. 

In conclusion, without interference, SEMcm outperforms the LSE approach in estimating the mean ($10\log{}_{10}(\Omega m)$), while it is comparable to ML in estimating $m$.
\begin{figure}[t] 
\begin{center}
\includegraphics[width=3.4in]{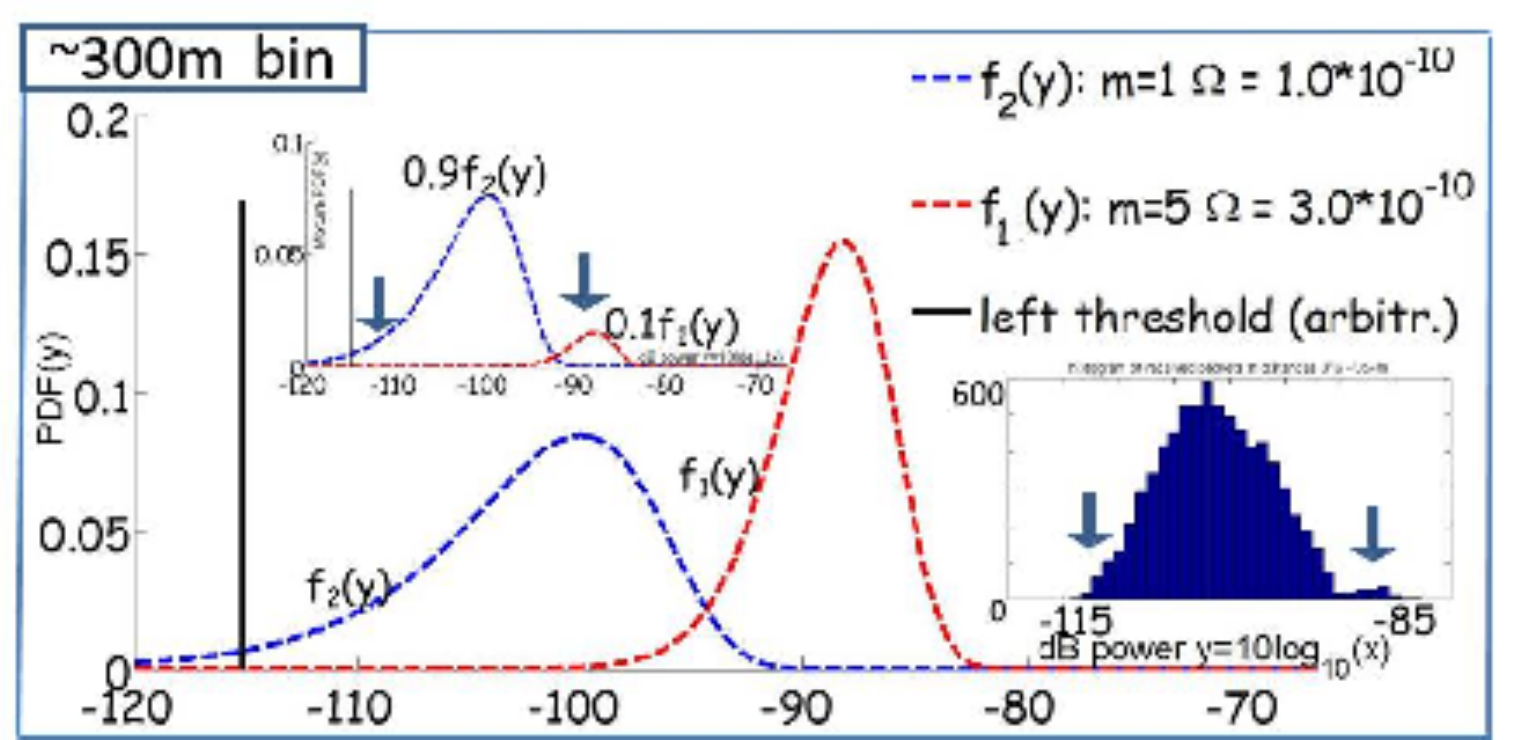}\vspace{-1mm}
\caption{log-Gamma component PDFs ($f_1$ and $f_2$), based on the SEMcm estimates, given data from the 3rd part in Fig. 1 for a given distance bin.  Mixing coefficients are found to be 0.1 and 0.9: mixture PDF with these $\alpha_i$ is shown as insert, along with the RSSI histogram of that distance bin.  The arrows point to the similarity of the estimated PDF shapes and empirical data.} \vspace{-3mm}  \label{fig:GammaCISS}
\end{center}
\end{figure}
Finally,  we present Fig.~\ref{fig:GammaCISS} which is based on the data featured in Fig 1. Apart from show-casing the notion of dual mixture and censoring, this figure affirms the censored mixture approach, as it illustrates a good match between the SEM-reconstructed PDF of the data featured in Fig 1, and its empirical distribution. Observe that the points left of the black vertical line around $-115 dBm$ represent censored samples (i.e, $c_L=-115$) .
\section{Conclusion}
Our main contribution is a novel model of interference affected RSSI samples, presented as censored mixture of Gamma PDFs, based on the insight from data collected for varying interference levels (see Fig. 1). Also, we applied a flavor of EM algorithm which not only mechanizies the computation of the parameters' ML estimates for our complex statistical model of {\em incomplete non-Gaussian mixed data} \cite{DempsterLairdRubin,LeeScott}, but also utilizes stochastic randomization to avoid strong dependence on its starting position, convergence to a saddle point, and low convergence rate. A great property of this method is that it leverages the count of lost data, to improve estimates for small number of samples, which is especially important for online estimation based on crowd-sourced data. 

Our future work will explore online versions of EM algorithms \cite{onlineEMlatent} applied to our problem. Also, future work will address improvements for signal levels that are on average too close to interference levels, such as in cluster-overlap bins in Figures~\ref{fig:InterfMean}~and~\ref{fig:mixtM}. Although this is a common problem in data clustering, we believe that good predictive models for cluster overlaps could be developed based on signal samples in distance bins with good separation. 
\vspace{-0.1cm}
\bibliographystyle{IEEEtran}
\bibliography{SEMReferences}
\end{document}